\pdfoutput=1

\documentclass[11pt]{article}

\usepackage[preprint]{acl}

\usepackage{booktabs}
\usepackage{times}
\usepackage{latexsym}
\usepackage{multirow}
\usepackage{listings}
\usepackage{amsmath} 
\usepackage{pdflscape}
\usepackage{rotating}
\usepackage{xcolor}
\usepackage{array}
\usepackage{colortbl}
\usepackage{soul}
\usepackage{ulem}

\usepackage[T1]{fontenc}

\usepackage[utf8]{inputenc}

\usepackage{microtype}

\usepackage{inconsolata}

\usepackage{graphicx}
\usepackage{amssymb}
\def\Bx{\mbox{\boldmath $x$}}
\def\Bq{\mbox{\boldmath $q$}}

\title{Dynamic Few-Shot Learning for Knowledge Graph Question Answering}

\author{Jacopo D'Abramo \\
  University of Bologna, Bologna, Italy \\
  \texttt{jacopo.dabramo@studio.unibo.it} \And
  Andrea Zugarini \\
  Expert.ai, Siena, Italy \\
  \texttt{azugarini@expert.ai} \AND
  Paolo Torroni \\
  University of Bologna, Bologna, Italy \\
  \texttt{paolo.torroni@unibo.it} \\}

\begin{document}
\maketitle
\begin{abstract}
Large language models present opportunities for innovative Question Answering over Knowledge Graphs (KGQA). However, they are not inherently designed for query generation. To bridge this gap, solutions have been proposed that rely on fine-tuning or ad-hoc architectures, achieving good results but limited out-of-domain distribution generalization. In this study, we introduce a novel approach called Dynamic Few-Shot Learning (DFSL). DFSL integrates the efficiency of in-context learning and semantic similarity and provides a generally applicable solution for KGQA with state-of-the-art performance.
We run an extensive evaluation across multiple benchmark datasets and architecture configurations.
\end{abstract}

\section{Introduction}
The growth of the Semantic Web has led to the creation and storage of vast amounts of structured knowledge \citep{Hitzel:2021,Shadbolt:2006}, organized into massive Knowledge Graphs (KGs) such as Wikidata \citep{wikidata}, DBpedia \citep{Lehmann:2014}, and FreeBase \citep{Bollacker:2008}. The scale of these KGs, with over 109 million items in Wikidata alone,\footnote{\url{https://www.wikidata.org/wiki/Wikidata:Statistics}} has made extracting relevant information from them increasingly challenging. This led to the emergence of Knowledge Graph Question Answering (KGQA), whose goal is to answer natural language questions posed over KGs. 

A typical KGQA system consists of three main components: Entity Linking (EL), Relation Linking (RL), and Query Genration (QG). 
Starting from a natural language question $q$, EL and RL return a set of entities $\mathcal{E}_q$ and relations $\mathcal{R}_q$ therein. The QG module, crucially, takes $q$, $\mathcal{E}_q$ and $\mathcal{R}_q$ and generates a SPARQL query that produces the answer. 

This paper focuses on the QG component. State-of-the-art approaches to SPARQL query generation are based on fine-tuning language models like T5~\citep{TSET}, or ad-hoc architectures leveraging LLMs and dependency trees~\citep{Rony:2022}. Despite their success, such approaches have limited flexibility and scalability. Fine-tuning in particular 
may be computationally expensive and struggle with out-of-domain distributions.


This paper proposes a novel approach to KGQA, leveraging in-context learning with Large Language Models (LLMs). The main intuition is that \textit{a significant number of errors could be addressed by making better use of the examples in the training set}. 
Our methodology, termed Dynamic Few-Shot Learning (DFSL), leverages semantic search to retrieve similar questions from the training set and enrich the prompt accordingly. 

To evaluate the performance and robustness of DFSL, we run experiments on two widely-used Knowledge Bases, DBpedia and Wikidata, using four publicly available datasets: QALD-9, based on  DBpedia, and QALD-9 plus, QALD-10 and LC-QuAD 2.0, based on Wikidata. 
As backbones, we use three state-of-the-art LLMs: Mixtral 8x7B, Llama-3 70B, and CodeLlama 70B.
Our experimental results demonstrate that our model achieves new state-of-the-art results, with significant advantages in terms of speed and efficiency.
We also run ablation studies to gauge the effectiveness of the approach without gold information from the EL and RL modules.

Our main contributions are:
    (1) a novel approach to KGQA, called DFSL, that leverages semantic search for dynamic few-shot learning; (2) state-of-the-art results by a significant margin in most benchmarks; (3) an extensive evaluation and ablation study to investigate quantitatively and qualitatively the impact of hyperparameters, backbones, embedding methods, answer selection strategies and gold entity/relation information.

\begin{figure*}
    \centering
    \includegraphics[scale=0.71]{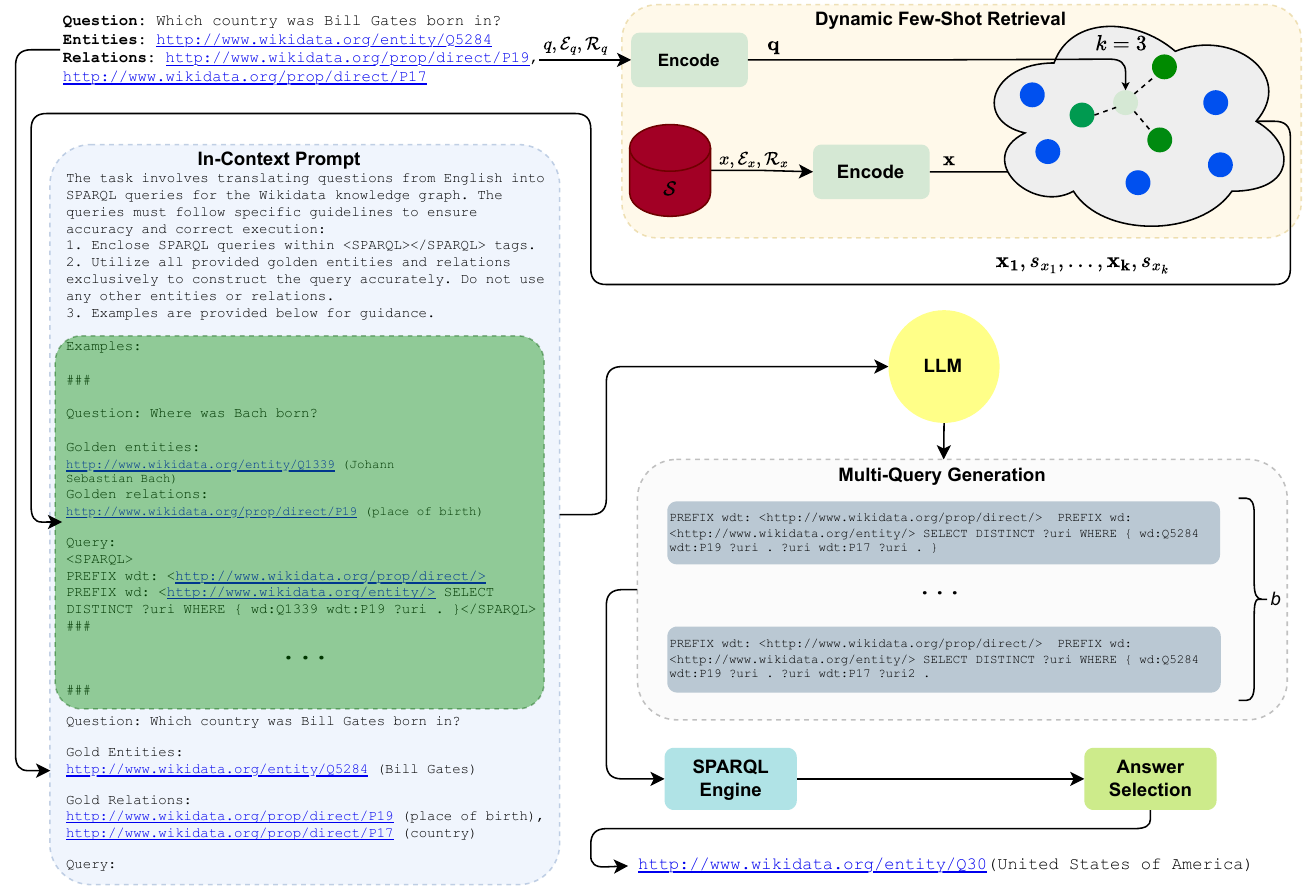}
    \caption{Sketch of DFSL. Given a question, its entities and its relations, $k$-most similar examples are retrieved from a text-to-SPARQL collection $\mathcal{S}$ and injected into the in-context prompt. Then, the LLM generates one or more queries that are all executed by a SPARQL engine. An answer selection strategy identifies which response to pick.}
    \label{fig:dfsl_sketch}
\end{figure*}
\section{Related work}
Early research in KG query generation was rule-based~\citep{guo:2005,Owens:2008}, template-based  \citep{ZENZ:2009,Unger:2012} or
search-based.
For example, \citet{Gorlitz:2012} developed a query generation heuristic to predict the final SPARQL representations by exhaustively checking all possible combinations of query patterns.
However, manual or semi-manual approaches hit scalability issues with KGs like WikiData and DBpedia. More recent approaches belong to two main streams: information-retrieval based methods and Text-to-SPARQL approaches.

%

\paragraph{Information Retrieval KGQA.} 
This family of methods 
involves the identification of sub-graphs relevant to $q$. Approaches include divide-and-conquer~\citep{kg-gpt}, fact retrieval based on linked entities~\citep{kaping}, more complex methods involving hops, relation predictions, and triple sampling~\citep{retrieve-rewrite}, or Evidence Pattern Retrieval (EPR) through structural dependency modeling~\citep{EPR}.

\paragraph{Text-to-SPARQL.} 
With the recent wave of decoder-based LLMs such as GPT \citep{GPT3}, Mixtral \citep{mixstral}, and LLamA \citep{llama}, generative AI was also used to translate $q$ into SPARQL queries. Notably, \citet{zou:2021} introduced a text-to-SPARQL model that leverages a relation-aware attention decoder and a pointer network encoder, incorporating three separate scaled dot-product attention mechanisms to generate SPARQL queries that capture entity, relation, and keyword representations.
\citet{Banerjee:2022} experimented with various models, including T5~\citep{Raffel:2020}, BART~\citep{BART}, and Pointer Generation Networks~\citep{PGN}, to explore their efficacy in KGQA tasks.
\citet{Rony:2022}'s SGPT employs a stack of transformer encoders to extract linguistic features from $q$ and GPT-2 as a decoder. However, this architecture is limited by its inability to capture connections among entities and relations in the underlying knowledge graph, leading to errors in generating triple sequences in the final SPARQL queries.
\citet{sparqlgen} presented a one-shot generative approach, where the prompt is augmented with a KG fragment required to construct the query and a question-subgraph query example.

Despite promising results, these architectures are prone to systematic errors. One such error, the so-called ``triple-flip", refers to the reversal of subject and object positions in the generated SPARQL triples, yielding wrong, often empty answers. \citet{TSET} addressed this issue by developing TSET, a fine-tuned T5 model with a pretraining stage called Triplet Structure Correction. This approach aims to deepen the model's understanding of triple order, establishing state-of-the-art performance on major KGQA datasets.

\paragraph{Example Selection in Few-Shot Learning.}
In-context learning (ICL) is a paradigm that leverages reasoning through analogies. A task description, question, and demonstration context are usually concatenated to create a prompt, which is then input into an LLM for prediction. Unlike fine-tuning, ICL performs predictions without gradient updates \citep{surveyICL}. Few-Shot Learning is a type of ICL where the demonstration context includes a few examples.
Owing to the effectiveness of ICL and the obvious advantage of building systems that don't need domain-specific training, a great deal of research and engineering efforts have been devoted to designing suitable prompts. ICL has been successfully applied to many NLP problems, including QA~\citep{chada2021fewshotqa,chen2023gotta} and KGQA~\citep{li2023few}. Some studies have also focused on the selection of in-context examples. In particular,
\citet{liu:2022} developed KATE, an unsupervised retriever that utilizes k-nearest neighbors and distance metrics (e.g., L2 distance and cosine similarity) to select in-context examples for tasks such as sentiment analysis, table-to-text generation, and question answering. 
\citet{levy:2023} explored the incorporation of diverse demonstrations into prompts for compositional semantic parsing task, demonstrating that such diversity leads to better structural coverage in target utterances.  \citet{kim:2022}  leveraged the generative capabilities of pre-trained language models to generate demonstrations for each class in downstream tasks, conditioned on test inputs and class information. \citet{gonen:2022} found that selecting examples based on perplexity, in particular lower perplexity, is an effective strategy. However, to the best of our knowledge, example selection has not yet been applied to KGQA.

\paragraph{Text-to-SQL.} A cognate domain, text-to-SQL, aims at the translation of natural language questions to SQL queries. There, \citet{rajkumar:2022} demonstrated a zero-shot and few-shot approach using simple prompts, achieving lower results compared to fine-tuned approaches with models such as GPT-3 \citep{GPT3} and CODEX \citep{codex}. \citet{nan:2023} introduced various strategies for selecting examples based on similarities/dissimilarities, selecting similar questions with the same difficulty level and dissimilar questions by using k-means clustering to obtain k diverse examples close to each centroid. More recently, \citet{act-sql} proposed an automatic chain-of-thought \citep{chainofthought} approach, where question slices are matched with all possible table and column names to identify the most relevant ones for a given question, using models such as GPT 3.5 and GPT 4. In spite of the similarities between text-to-SQL and text-to-SPARQL, the methods developed so far for the former are not applicable in the latter, where instead of a data model with a relatively small-sized set of tables and columns, the domain is modeled by a large-scale, semi-structured KG.




\section{Method}\label{sec:method}



Given a collection of natural language questions $\mathcal{Q}$ and a knowledge graph $\mathcal{G}:= (\mathcal{E},\mathcal{R}, \mathcal{F})$, where $\mathcal{E}$ are \textit{entities},  $\mathcal{R}$ are \textit{relations}, and $ \mathcal{F} \subseteq \mathcal{E} \times \mathcal{R} \times \mathcal{E}$ are \textit{facts}, 
KGQA is the problem of answering questions in $\mathcal{Q}$ based on $\mathcal{G}$. 
KGQA can be framed as a \textbf{text-to-SPARQL} task, where a question $q$ must be translated into a SPARQL query $s_q$ to be executed on $\mathcal{G}$ by a SPARQL engine, to return a  (possibly empty) answer $a$.
The entities and relations in $q$, denoted as $\mathcal{E}_q$ and $\mathcal{R}_q$, may be, and usually are, extracted from $q$ before generating $s_q$. 
Hence, query generation can be tackled as a conditional text generation problem given $q, \mathcal{E}_q$ and $\mathcal{R}_q$. 
Within the scope of ICL, $P_{\theta}$ is a pre-trained LLM and the conditional input $\mathcal{E}_q, \mathcal{R}_q$, $q$ is combined with other contextual information $C$, such as additional instructions, guidelines, constraints and demonstrations, all expressed via natural language text. Accordingly, the generated query is: 
\begin{equation}
    s_q = \arg\max_{s} P_{\theta}(s | C, \mathcal{E}_q, \mathcal{R}_q, q).\label{eq:icl_sparql}
\end{equation}



\begin{table*}
\centering
\small
\begin{tabular}{lccccc}
\toprule
\textbf{Approach} & \textbf{Backbone} & \textbf{QALD-9 Plus} & \textbf{QALD-10} & \textbf{LC-QUAD 2.0} & \textbf{QALD-9 DB} \\
\midrule
Zero-shot Learning & & 49.90 & 33.76 & 40.66 & 65.73 \\
Few-shot Learning & Mixtral 7x8 & 54.80 (\textbf{+4.90}) & 50.26 (\textbf{+16.50}) & 61.04 (\textbf{+20.38}) & 63.86 (\textbf{-1.87}) \\
DFSL & & 71.75 (\textbf{+21.85}) & 49.90 (\textbf{+16.14}) & 81.81 (\textbf{+41.15}) & 72.74 (\textbf{+7.01}) \\
\midrule
Zero-shot Learning & & 63.01 & \textbf{58.31} & 54.21 & 70.49 \\
Few-shot Learning & Llama-3 70B & 67.69 (\textbf{+4.68}) & 51.28 (\textbf{-7.03}) & 68.52 (\textbf{+14.31}) & 68.84 (\textbf{-1.65}) \\
DFSL & & 73.60 (\textbf{+10.59}) & 56.59 (\textbf{-1.72}) & 81.93 (\textbf{+27.72}) & 72.66 (\textbf{+2.17}) \\
\midrule
Zero-shot Learning & & 45.94 & 33.36 & 38.40 & 66.43 \\
Few-shot Learning & CodeLlama 70B & 64.49 (\textbf{+18.55}) & 57.38 (\textbf{+24.02}) & 64.46 (\textbf{+26.06}) & 72.67 (\textbf{+6.24}) \\
DFSL & & \textbf{76.59} (\textbf{+30.65}) & 57.69 (\textbf{+24.33}) & \textbf{85.45} (\textbf{+47.05}) & \textbf{75.14} (\textbf{+8.71}) \\
\bottomrule
\end{tabular}
\caption{Comparison between zero-shot, few-shot and DFSL with different backbones. Absolute F1 gains with respect to the naive zero-shot approach are reported between parenthesis.}
\label{tab:backbones_results} 
\end{table*}

\begin{table*}
    \centering 
    \begin{tabular}{lcccc} 
        \toprule 
        \textbf{Approach} & \textbf{QALD-9 Plus} & \textbf{QALD-10} & \textbf{LC-QUAD 2.0} & \textbf{QALD-9 DB}\\
        \midrule 
        DFSL & 76.59 & 57.69 & 85.45 & 75.14  \\   \midrule
        $\text{DFSL-MQP}_{\text{LS}}$ & 73.67 & 58.85 & 85.06 & 73.25\\
        $\text{DFSL-MQP}_{\text{FS}}$ & 74.40 & 58.34 & 85.38 &  73.92 \\ \midrule
        $\text{DFSL-MQ}_{\text{LS}}$   & 83.21  & 60.48       & 85.54   & 72.06   \\
        $\text{DFSL-MQ}_{\text{FS}}$ & \textbf{84.45} (\textbf{+7.86})  & \textbf{62.20}  (\textbf{+4.51}) & \textbf{89.10} (\textbf{+3.65}) & \textbf{77.89} (\textbf{+2.75})\\
        \bottomrule
    \end{tabular}
    \caption{Multi-query Generation: comparing DFSL-MQ with DFSL and Multi-query prompting baselines. Absolute F1 gains with respect to DFSL are reported for the best performing configuration.}
    \label{tab:mq_results} 
\end{table*}

\subsection{Dynamic Few-Shot Retrieval}\label{subsec:dynamic_fs_retrieval}
In few-shot ICL, the choice of demonstrations to inject in the prompt can significantly affect performance. Usually, few-shot examples are predetermined representative instances of the task, hand-picked during prompt design.
Conversely, we aim to retrieve good examples dynamically, based on their relevance to the input question. Inspired by \citet{liu:2022}, we 
adopt a retrieval approach based on the similarity between a question $q$ and a set of previously answered text-to-SPARQL examples collected in a storage $\mathcal{S}$ (see Figure~\ref{fig:dfsl_sketch}), where each example is a tuple including a question $x$, its entities $\mathcal{E}_{x}$ and relations $\mathcal{R}_{x}$, and the associated SPARQL query $s_{x}$.
The question, its entities and relations $\langle q, \mathcal{E}_q, \mathcal{R}_q\rangle$ are mapped onto a vector representation $e_q \in \mathbb{R}^d$ using a sentence encoder. To properly feed such information to an encoder-only LM, we concatenate question, entities and relations in a single input sequence $\Bq := [q, \mathcal{E}_q,\mathcal{R}_q]$.
Likewise, we encode each example $x \in \mathcal{S}$ into a vector $e_x \in \mathbb{R}^d$ and then compute the similarity between the target question and the storage: 
\begin{equation}
    score(\Bq, \Bx) = sim(e_q, e_x), \forall x \in \mathcal{S},
\end{equation}
where the $sim$ is a similarity function.
Based on such a scoring, we retrieve the $k$-most similar examples $\mathcal{S}$ and include them as demonstrations in the in-context prompt.

\subsection{In-Context Prompt} 
The in-context prompt has three parts. The first is the task description, instructing the LLM with a numbered list of guidelines on the output format and on the available information. 
The second, highlighted in Figure~\ref{fig:dfsl_sketch} with a green block, contains the $k$ retrieved demonstrations. Each demonstration consists of a question, its entities and relations, denoted as \textit{gold} entities/relations, all paired with their SPARQL query delimited by \texttt{<SPARQL></SPARQL>} tags. The \textbf{\#\#\#} symbol delimits each single example.
The final part is the input question, associated with its gold entities and relations.
The answer returned by the LLM prompted as such is then parsed to extract the generated text enclosed in \texttt{<SPARQL></SPARQL>} tags. The resulting query $s_q$ is executed by a SPARQL engine on $\mathcal{G}$ to yield the answer to $q$. 
We call our approach Dynamic Few-Shot Learning (DFSL).

\subsection{Multi-Query Generation}
A typical challenge faced by LLMs in SPARQL query generation is the understanding of what is the subject and what is the object of a relation, an information the model does not have.  This problem is called triple-flip error~\citep{TSET}. LLMs often end up in swapping the subject with the object in the query, almost choosing one way or the other randomly. Thanks to DFSL, this issue may be alleviated whenever there are similar demonstrations in the in-context prompt that clarify the subject-object roles. To further reduce triple-flip errors, we propose the generation of multiple SPARQL queries by retaining all the final hypotheses generated during beam search. 
The model uncertainty in placing subject and object is likely to be reflected in the beam search exploration. Intuitively, both triple-ordering hypotheses are considered plausible by the model. Thus, instead of just returning the most probable sequence $s$ according to Equation~\ref{eq:icl_sparql}, we keep the whole $b$ queries $\{s_{q,1}, \ldots, s_{q,b}\}$ formulated by beam search. We use DFSL-MQ to denote such a multi-query extension of DFSL.

\paragraph{Answer Selection.} Executing multiple queries inevitably leads to multiple possible answers. Therefore, we must define an answer selection criterion. We designed two heuristics: Largest Set (LS) and First Set (FS). LS executes all the $b$ queries, obtaining with each query $s_{q,j}$ a (possibly empty) answer set $\mathcal{A}_j$. LS then selects, among $\{ \mathcal{A}_1, \ldots, \mathcal{A}_b\}$, the largest one\footnote{In case of ties, we take the first largest set.}, i.e: \begin{equation*}
    \mathcal{A} = \arg\max_{\mathcal{A}_i}(|\mathcal{A}_1|,\ldots,|\mathcal{A}_b|),
\end{equation*}
\noindent the rationale being that incorrect candidates will likely have empty results. However, LS can be misled into selecting answers from under-constrained queries that return many irrelevant instances.
FS adheres to the natural beams ordering by selecting the first query that yields a non-empty answer set.

\begin{figure*}[h]
    \centering
    \includegraphics[scale=0.35]{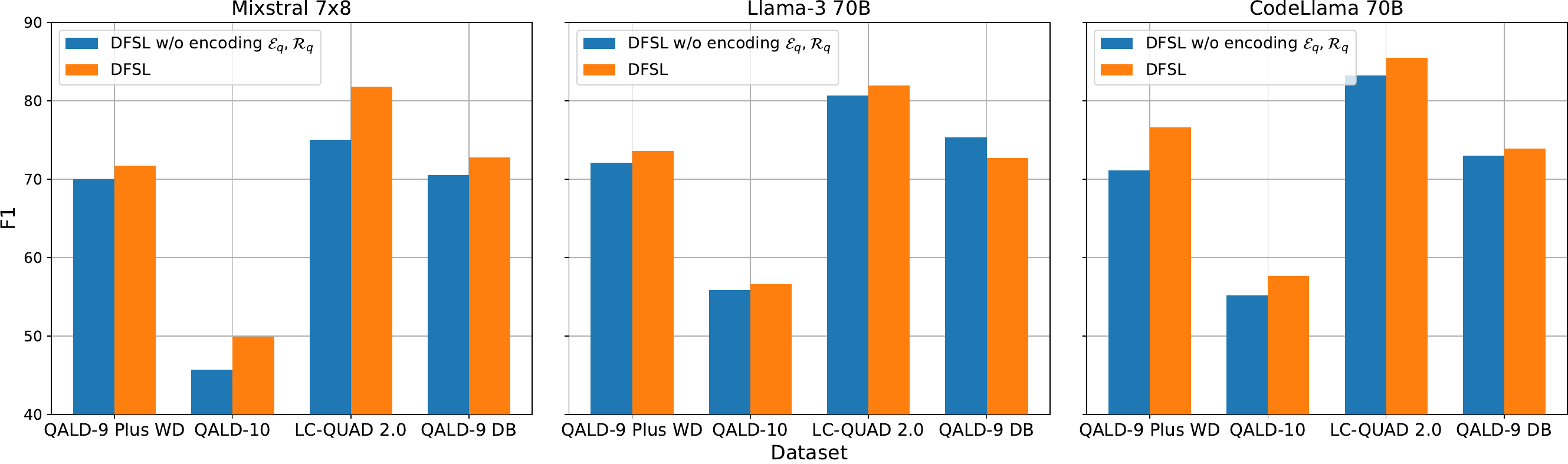}
    \caption{Comparison of Embeddings: DFSL (in orange) encoding that incorporates question, entities and relations versus an embedding solely based on the question $q$ (in blue).}
    \label{fig:different_embeddings}
\end{figure*}

\section{Experiments}
In this section, we aim to study the effects of each component involved in our DFSL approach. We evaluate DFSL and its extension DFSL-MQ on four KGQA datasets. In our investigation, we consider different backbones and we compare with multiple baselines and state-of-the-art solutions.

\subsection{Datasets}
To assess the flexibility and robustness of our approach, we evaluate it on four heterogeneous KGQA benchmarks based on two different Knowledge Graphs (Wikidata, DBpedia).

\paragraph{QALD-9 DB.} QALD-9~\citep{ngomo20189th} is a dataset from the Question Answering over Linked Data (QALD) challenge series. It comprises 408 training questions and 150 test questions. Unlike the other KGQA benchmarks, the SPARQL queries are meant for a DBpedia Knowledge Graph. We refer to it as QALD-9 DB to emphasize that.

\paragraph{QALD-9 plus.} QALD-9 plus 
extends QALD-9 on new languages and transfers SPARQL queries from DBpedia to Wikidata.  Although some queries were not portable to Wikidata due to the absence of corresponding information, it still comprises 371 training questions and 136 test questions.  In our experiments, we only consider English questions.

\paragraph{QALD-10.} QALD-10~\citep{qald10} is the latest dataset in the QALD  series, designed to increase the complexity of gold SPARQL queries. It consists of 412 training questions extracted from QALD-9 plus Wikidata. The test set was created from scratch, comprising 394 test questions that express real-world information needs. Test questions significantly differ from those in training.

\paragraph{LC-QuAD 2.0.} LC-QuAD 2.0 ~\citep{LC-QUAD2.0} is a large-scale dataset grounded on Wikidata. It consists of 30,226 simple and complex questions: 24,180 in training, and 6,046 in test. Questions are diverse. They include single- and multi-fact, boolean, count, and other query types. LC-QuAD 2.0 allows us to gauge the DFSL performance against a large text-to-SPARQL storage.

\subsection{Backbones}
\paragraph{Mixtral 8x7B.} Based on the Sparse Mixture of Experts (SMoE) architecture~\citep{SMoE}, Mixtral 8x7B~\citep{mixstral} is 
a 46.7B parameters model. Among the backbones adopted in this paper, Mixtral is the smallest. Moreover, thanks to the characteristics of its SMoE architecture, less than 13B are active at each inference step, making Mixtral particularly efficient.

\paragraph{Llama-3 70B.} Built upon the Llama architecture~\citep{llama}, Llama-3 70B has been trained on 15T tokens, a 650\% increase from its predecessor, Llama 2. At the moment we are writing, Llama-3 70B is one of the best-performing open-weights LLMs available.

\paragraph{CodeLlama 70B.} Initialized from Llama2 70B, CodeLlama~\citep{CodeLlama2} is a specialized version fine-tuned on 1T tokens of code-heavy data. Therefore, we expect CodeLlama to be particularly suitable for SPARQL query generation.

\subsection{Baselines}\label{sec:baselines}
\paragraph{Plain Question.} This is a naive baseline where we feed an LLM only with the task description and the question $q$. Without in-context examples nor any entity or relation associated with $q$, the LLM can only rely on its parameter memory.

\paragraph{Zero-Shot Learning.} Here we do not provide any demonstrative example in the prompt. However, unlike the plain question baseline, we do inject golden entities and relations into the prompt. With reference to Figure~\ref{fig:dfsl_sketch}, the In-Context prompt remains the same but without the green-like block containing the demonstrations.

\paragraph{Few-Shot Learning.}  The prompt is filled with a single set of $k$ manually selected examples, used for all the questions in the test set. The examples were chosen to maximize diversity and cover different kinds of queries\footnote{The chosen examples and more details are provided in Appendix~\ref{app:fs_examples}.}.

\paragraph{Multi Query Prompting (DFSL-MQP).} As an alternative to our proposed multi-query generation (DFSL-MQ), this baseline consists in a naive multi-query prompting strategy. Essentially, we ask the model to generate more queries to answer the question. To ease the creation of inverted subject-object queries that can solve triple-flip errors, we extend the prompt to explicitly ask the model to produce this kind of SPARQL queries. Answer selection uses LS and FS heuristics, like with DFSL-MQ.

\subsection{Experimental Setup}
\textbf{Implementation.}
In our experiments, the training set of each dataset serves as 
storage for the retrieval of the $k$ most similar examples (see the next paragraph for details on $k$ tuning) with DFSL. 
Examples are encoded with a sentence transformer\footnote{\url{https://www.sbert.net/index.html}}, all-mpnet-base-v2\footnote{\url{https://huggingface.co/sentence-transformers/all-mpnet-base-v2}}, and $sim$ is defined as the cosine similarity.
Inference is performed via beam search in both DFSL, where $b$ is set to 3, and DFSL-MQ, with $b$ set to 10. All the experiments were run on a cluster of 4 NVIDIA A100 GPUs.

\begin{figure}[!ht]
    \centering
    \includegraphics[scale=0.35]{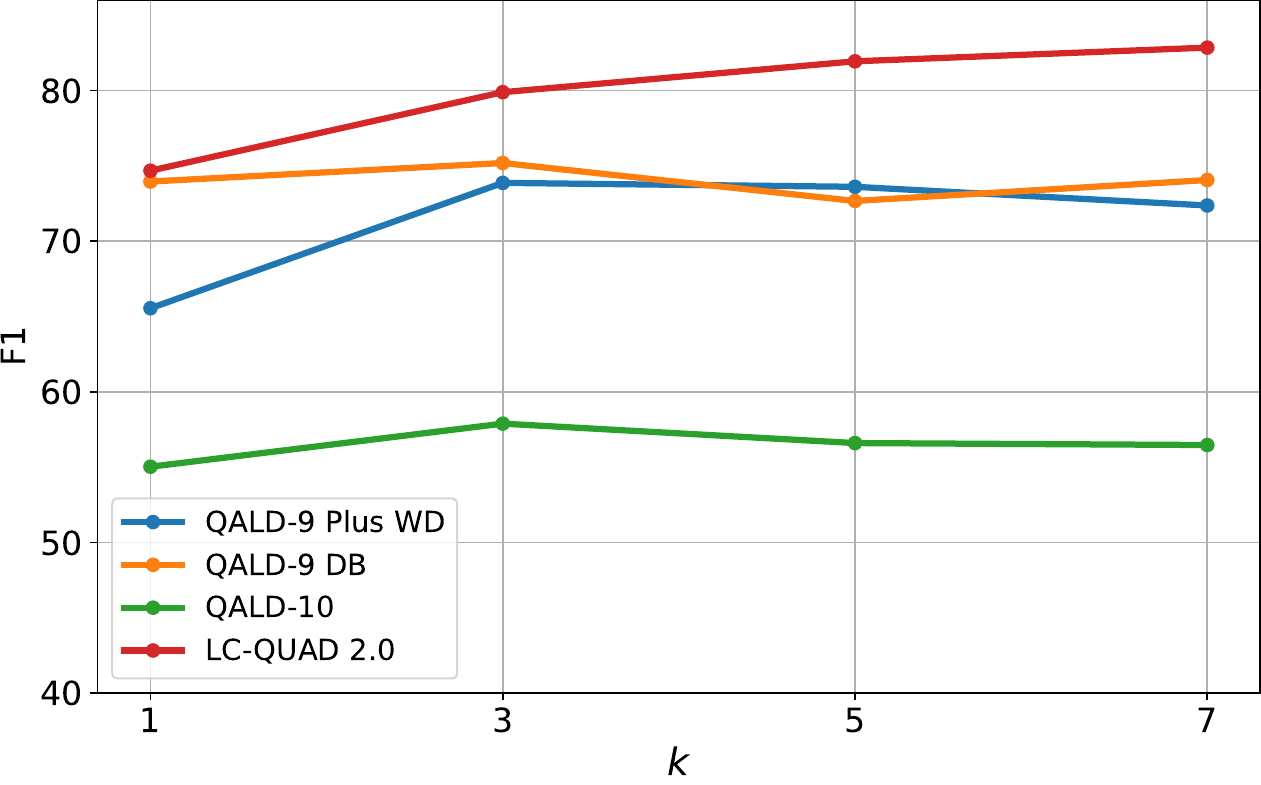}
    \caption{Impact of the number of in-context examples on the four benchmarks.}
    \label{fig:top_k_llama3}
\end{figure}

\paragraph{Number of Few-shot Examples.}
We first analyzed how the number of few-shot examples $k$ retrieved by DFSL affects the performance. We chose among $k=\{1,3,5,7\}$ and evaluated DFSL with Llama 3 70B backbone on the four datasets. The results shown in Figure \ref{fig:top_k_llama3} suggest that values of $k$ greater than one perform comparably well on smaller benchmarks, while on LC-QUAD 2.0, where there are about 25 thousands examples as storage, increasing $k$ seems to be beneficial. This may be due to the increased likelihood of finding similar examples in larger datasets as $k$ grows. We set $k=5$ for all the forthcoming experiments, which is a good trade-off across all the datasets. 


\begin{table*}
    \centering
    \small
    \begin{tabular}{lcccc} 
        \toprule 
        \textbf{Approach} & \textbf{QALD-9 Plus} &\textbf{QALD-10} & \textbf{LC-QUAD 2.0} & \textbf{QALD-9 DB} \\
        \midrule 
        Plain Question & 0.08 & 0.02 & 12.00 & 16.42\\ 
        BART     \citep{Banerjee:2022}      &   -        &  -          & 64.00 & - \\
        PGN-BERT-BERT \citep{Banerjee:2022} &   -        &  -          & 86.00 & - \\
        SGPT \citep{Rony:2022}              &   -        &  -          & 89.04 & 67.82 \\
        TSET-small \citep{TSET}             & 72.86      & 47.15       & 94.00       & -  \\
        TSET-base \citep{TSET}        & 75.85       & 51.37       & \textbf{95.00}  & -        \\
        \midrule
        Zero-shot Learning  & 45.94 & 33.36 &  38.40 & 66.43  \\
        Few-shot  Learning   & 64.49 & 57.38 &  64.46 &  72.67 \\

        DFSL & 76.59 & 57.69 & 85.45 & 75.14\\   
        DFSL-MQ beam FS  & \textbf{84.45} (\textbf{+8.60})   & \textbf{62.20} (\textbf{+10.83})       & 89.10 (\textbf{-5.90}) & \textbf{77.89} (\textbf{+10.07})\\
        \bottomrule
    \end{tabular}
    \caption{DFSL and ICL approaches vs state-of-the-art fine-tuned models.}
    \label{tab:sota} 
\end{table*}

\paragraph{Prompt.}
The prompt illustrated in Figure~\ref{fig:dfsl_sketch} constitutes the default template in our experimentations. However, slight variations
are required in certain cases. For example, when running experiments on DBpedia knowledge graph, we replace the Wikidata reference with DBpedia in the first text segment. When we study the absence of gold information instead, we remove all the references to gold entities/relations (according to the ablation) from the entire prompt.
There are no differences in the prompts layout when running few-shot-learning baseline experiments. In zero-shot learning, only the in-context examples any reference to them are removed, all else being equal.

\paragraph{Evaluation metric.} We follow a standard  F1 score evaluation in KGQA benchmarks. The F1 is computed between the answer set returned by the target SPARQL query and the predicted one. 
When both the queries return an empty set, we assign an F1 score of 1. 
The F1 scores of all the examples are then averaged.


\subsection{Results}

\paragraph{Impact of Dynamic Examples.} 
To measure the importance of retrieving few-shot examples dynamically, we compare DFSL on different backbones against Zero-Shot and Few-Shot Learning baselines. Results are outlined in Table~\ref{tab:backbones_results}. 

In terms of backbones, Llama 3 consistently outperforms both Mixtral and CodeLlama in zero-shot learning scenario, whereas in few-shot, results are generally comparable between Llama-3 and CodeLlama. Such a strong Llama 3 zero-shot performance may be caused by some sort of data contamination, however we leave such an investigation for future works. 

Both few-shot learning and DFSL generally yield substantial gains with respect to zero-shot baseline on all the backbones and datasets. An exception occurs in QALD-10 with Llama-3. Notably, when comparing DFSL and Few-shot Learning baseline, we can see our approach improving F1 scores by a large margin in LC-QUAD 2.0, QALD-9 Plus and QALD-9 DB, with F1 increasing up to 21 absolute points\footnote{Some qualitative examples illustrate the benefits of DFSL over few-shot learning in Appendix~\ref{app:qualitative_analysis} 
 (see Table~\ref{tab:Few_shot_errors}).}. In QALD-10 instead, where the test set has a different distribution from its training, there are no significant differences between DFSL and the standard few-shot learning approach. Indeed, an approach like DFSL brings little benefits when the storage only contains unrelated examples. 

Overall, DFSL with CodeLlama  achieved the greatest performance with respect to all the other configurations. Therefore, we adopt CodeLlama as our backbone in the following DFSL experiments.


\paragraph{Impact of Multi-Query Generation.}
Here we investigate DFSL-MQ, the multi-query approach extending DFSL.
We evaluate both answer selection strategies, LS and FS, and compare them against the plain DFSL and the multi-query prompting baseline described in Section~\ref{sec:baselines}. All the results are outlined in Table~\ref{tab:mq_results}. 

Having multiple queries is not necessarily beneficial. Indeed, the multi-query prompting baseline under-performs in three datasets out of four with respect to (single query) DFSL, regardless of the answer selection method adopted. DFSL-MQ instead proves to be generally beneficial. Both Largest Set and First Set heuristics are effective when the hypotheses come from the beams. Furthermore, FS consistently outperforms LS, even by substantial margins in QALD-9 DB.

\paragraph{In-context Learning vs Fine-tuning.}
Up to this point, we have assessed DFSL in the scope of In-Context Learning approaches. In Table~\ref{tab:sota} instead, we compare our approach against state-of-the-art models trained and/or fine-tuned for specific downstream KGQA datasets. Without any training, DFSL-MQ outperforms current state-of-the-art approaches in three out of four benchmarks, namely QALD-9 Plus, QALD-10 and QALD-9 DB, even with the single query DFSL setup. DFSL-MQ does not obtain state-of-the-art results in LC-QUAD 2.0, the dataset most affected by triple-flip errors. This means that multi-query generation only alleviates the issue, but the problem still remains.

\subsection{Ablation studies}

\paragraph{Different Example Encoding.}
As described in Section~\ref{subsec:dynamic_fs_retrieval}, to compute the embeddings we concatenated the textual input made of the question and its list of entities and relations. Here, we gauge the impact of this additional information on DFSL performance. In Figure~\ref{fig:different_embeddings} we compare DFSL, with a variant where we only embed the natural language question $q$, without any additional data concatenated. The evaluation carried out in all the benchmarks and with all the backbones, demonstrates that such information improves the quality of the generated queries.

\begin{table}[ht]
    \centering
    \begin{tabular}{lc}
        \toprule
        \textbf{Approach} & \textbf{QALD-9 DB} \\ \midrule
        Plain Question & 16.42 \\ \midrule
        DFSL & 75.14 \\ \midrule
         DFSL w/o $\mathcal{R}_q$  & 56.62 (-18.56) \\
         DFSL w/o $\mathcal{E}_q$  & 60.92  (-14.22) \\
         DFSL w/o $\mathcal{E}_q, \mathcal{R}_q$  & 49.59 (-25.55) \\
    \bottomrule
    \end{tabular}
    \caption{DFSL in the absence of entities and/or relations.}
    \label{tab:ablation_gold_no_gold}
\end{table}

\paragraph{Absence of gold information.}
In KGQA, text-to-SPARQL generation usually relies not only on the question itself, but also on entities and relations associated to it. 
Here we assess DFSL  when either the entities $\mathcal{E}_q$ or the relations $\mathcal{R}_q$, or both are missing.
The information is removed throughout the entire process. For example, when removing entities, we discard them from both the storage and the prompt. Even the embeddings for the retrieval are computed by encoding an input without any entity concatenated in $\Bq$, i.e. becoming $\Bq = [q, \mathcal{R}_q]$. 
We report this on QALD-9 DB dataset. By observing the results outlined in Table~\ref{tab:ablation_gold_no_gold}, it is clear that, without full knowledge of the entities and the relations required for generating the query, the LLM performance drops significantly. Nonetheless, even in the case where no information is given (DFSL w/o $\mathcal{E}_q, \mathcal{R}_q$), the presence of dynamic demonstrations is essential, yielding a 33+ absolute F1 increase compared to plain question baseline.

\section{Conclusion}
In this paper, we introduced DFSL, a novel approach to Knowledge Graph Question Answering. This method leverages semantic search to dynamically retrieve relevant examples from the training set, enriching the prompt for LLMs to improve the generation of SPARQL queries. We conducted comprehensive experiments on four publicly available datasets based on two widely-used KBs, DBpedia and Wikidata. By employing three different state-of-the-art LLMs as backbones, we demonstrated that DFSL achieves superior performance compared to both standard in-context learning techniques and state-of-the-art models fine-tuned on the downstream task.
We further conducted an extensive evaluation of DFSL through ablation studies to measure the impact of hyper-parameters, different backbones, embedding methods, answer selection strategies, and the inclusion or exclusion of entities and relations information associated to a question.
The code will be released publicly upon acceptance of the paper.
In the future, we plan to study the effectiveness of DFSL in cognate domains like text-to-SQL.

\section*{Limitations}
We recognize some limitations in our work. 
Our experiments are all on English-based datasets, where notoriously LLMs are better performing. Moreover, the massive pre-training of those LLMs on a vast portion of the Web, may expose those models to unintended data contamination.
Experiments only focused on LLMs with large number of parameters, without investigating the behaviour of smaller models. To encode examples, we limited the investigation to what kind of text to encode (just the question, or the question and its entities and relations), without exploring different embedding models, similarity criteria or other input concatenation strategies.
We leave these investigations to future work.

\bibliography{acl_latex}

\onecolumn
\appendix
\section{Qualitative Analysis}\label{app:qualitative_analysis}
In this appendix we provide some qualitative analyses of DFSL and DFSL-MQ.  
First of all, we report some examples in Table~\ref{tab:Few_shot_errors} that highlight the benefits from introducing similar examples with DFSL with respect to standard few-shot learning approach. 
Then, we show some examples in Table~\ref{tab:DFSL_errors} where the multi-query approach solves triple-flip errors.
In Table~\ref{tab:LS_errors} instead, we showcase errors caused by employing LS answer selection heuristic. Notably, by choosing larger sets, LS sometimes selects queries that are often relegated to latter positions in the beam hypotheses, which tend to be more general, thus more prone to returning a greater number of results.

\definecolor{greenforest}{RGB}{124, 176, 132}
\definecolor{softred}{RGB}{255, 204, 203}
\definecolor{red}{RGB}{255, 0, 0}

\newcolumntype{P}[1]{>{\arraybackslash\vspace{0pt}}p{#1}}

\begin{table*}[ht!]
  \centering
  \small
  \resizebox{\textwidth}{!}{
  \begin{tabular}{|P{3cm}|P{3.5cm}|P{1.25cm}|P{4.5cm}|P{0.75cm}|}
    \hline
    \textbf{Question} & \textbf{Target Query} & \textbf{Answer Selection} & \textbf{Predicted Query} & \textbf{Beam}\\
    \hline
    What is manufactured NEC PC-9800 series whose sector is electronics? &
   \texttt{SELECT ?answer WHERE \{ wd:Q183505 wdt:P176 ?answer .  ?answer wdt:P452 wd:Q11650 \}} &
   \multirow{2}{*}{LS} & 
   \cellcolor{softred}\texttt{SELECT ?answer WHERE \{ wd:Q183505 wdt:P176 ?X .  ?X wdt:P452 wd:Q11650 . ?X wdt:P31 ?answer\}} & 
   7 \\
   \cline{3-5} 
   &  & 
   FS & 
   \cellcolor{greenforest}\texttt{select distinct ?obj where \{ wd:Q183505 wdt:P176 ?obj . ?obj wdt:P452 wd:Q11650 \}} & 
   4  \\
   \hline

   What is the enthalpy of vaporization for potassium hydroxide? &
   \texttt{select distinct ?answer where \{ wd:Q132298 wdt:P2116 ?answer\}} &
   \multirow{2}{*}{LS} & 
   \cellcolor{softred}\texttt{ASK WHERE \{ wd:Q14982 wdt:P2116 ?obj filter(?obj = 37400) \}} & 
   3 \\
   \cline{3-5}
   &  & 
   FS & 
   \cellcolor{greenforest}\texttt{select distinct ?answer where \{ wd:Q132298 wdt:P2116 ?answer\}} & 
   1  \\
   \hline
   
   Which is the rural city of Fukushim? &
   \texttt{SELECT ?answer WHERE \{ wd:Q161176 wdt:P131 ?answer . ?answer wdt:P150 wd:Q1347240\}} &
   \multirow{2}{*}{LS} & 
   \cellcolor{softred}\texttt{SELECT ?answer WHERE \{ wd:Q161176 wdt:P131 ?X . ?X wdt:P150 ?answer\}} & 
   3 \\
   \cline{3-5}
   &  & 
   FS & 
   \cellcolor{greenforest}\texttt{SELECT ?answer WHERE \{ wd:Q161176 wdt:P131 ?answer . ?answer wdt:P150 wd:Q1347240\}} & 
   1  \\
\hline
  \end{tabular}
  }
  \caption{
Qualitative comparison between different answer selection strategies in DFSL-MQ.}\label{tab:LS_errors}
\end{table*}
\newcommand{\hlgreenforest}[1]{{\sethlcolor{greenforest}\hl{#1}}}
\newcommand{\hlred}[1]{{\sethlcolor{softred}\hl{#1}}}
\begin{table*}[ht!]
  \centering
  \resizebox{\textwidth}{!}{
  \small
  \begin{tabular}{|P{2.5cm}|P{3cm}|P{1.5cm}|P{4cm}|P{3cm}|}
    \hline
    \textbf{Question} & \textbf{Target Query} & \textbf{Approach} & \textbf{Predicted Query} & \textbf{Similar In-context Example}\\
    \hline

    Who is the daughter of Robert Kennedy married to? &
   \texttt{SELECT DISTINCT ?uri WHERE \{ wd:Q25310 wdt:P40 ?daughter . ?daughter  wdt:P21 wd:Q6581072 . ?daughter  wdt:P26 ?uri .\}} &
   \multirow{2}{*}{Few-Shot} 
   & \texttt{SELECT DISTINCT ?uri WHERE \{ \hlred{?uri wdt:P40 wd:Q25310} ; wdt:P21 wd:Q6581072 ; \hlred{wdt:P26 ?spouse .}} 
   &  - \\
   \cline{3-5} 
   &  & 
   DFSL & 
   \texttt{SELECT DISTINCT ?uri WHERE \{ \hlgreenforest{wd:Q25310 wdt:P40 ?child} . ?child wdt:P21 wd:Q6581072 . \hlgreenforest{?child wdt:P26 ?uri }. \}} & 
   \texttt{SELECT DISTINCT ?uri WHERE \{ \hlgreenforest{wd:Q43247 wdt:P40 ?child . ?child wdt:P26 ?uri .} \}}  \\
   \hline

   Which countries are connected by the Rhine? &
   \texttt{SELECT DISTINCT ?uri WHERE \{ wd:Q584  wdt:P17 ?uri . ?uri wdt:P31 wd:Q6256 . \}}  &
   \multirow{2}{*}{Few-Shot} 
   & \texttt{SELECT DISTINCT ?uri WHERE \{ ?uri wdt:P31 wd:Q6256 ; \hlred{wdt:P17 wd:Q584 .}\}} 
   & - \\
   \cline{3-5}
   &  & 
   DFSL & 
   \texttt{SELECT DISTINCT ?uri WHERE \{ \hlgreenforest{wd:Q584 wdt:P17 ?uri} . ?uri wdt:P31 wd:Q6256 . \}} & 
   SELECT DISTINCT ?res WHERE \{ \hlgreenforest{wd:Q3392} wdt:P885/\hlgreenforest{wdt:P17 ?res .} \}  \\
   \hline

   Give me the capitals of all countries in Africa. &
   \texttt{SELECT DISTINCT ?uri WHERE \{ ?country wdt:P31 wd:Q6256 . ?country  wdt:P30 wd:Q15 . ?country wdt:P36 ?uri . \}}  &
   \multirow{2}{*}{Few-Shot} 
   & \texttt{SELECT DISTINCT ?uri WHERE \{ ?uri wdt:P31 wd:Q6256 ; wdt:P30 wd:Q15 ; \hlred{wdt:P36 ?capital .} \}} 
   & - \\
   \cline{3-5}
   &  & 
   DFSL & 
   \texttt{SELECT DISTINCT ?uri WHERE \{ ?country wdt:P31 wd:Q6256 . ?country wdt:P30 wd:Q15 . \hlgreenforest{?country wdt:P36 ?uri} . \}} & 
   SELECT DISTINCT ?uri WHERE \{ wd:Q5451 wdt:P17 ?country . \hlgreenforest{?country wdt:P36 ?uri} . \}  \\

    \hline

  \end{tabular}
  }
  \caption{A qualitative comparison between DFSL and Few-shot Learning.  Thanks to the similar in-context examples retrieved with DFSL, the LLMs generates the correct query.
 }\label{tab:Few_shot_errors}
\end{table*}

\begin{table*}[ht!]
  \centering
  \small
  \resizebox{\textwidth}{!}{
  \begin{tabular}{|P{4cm}|P{5cm}|P{1.5cm}|P{5.5cm}|}
    \hline
    \textbf{Question} & \textbf{Target Query} & \textbf{Approach} & \textbf{Predicted Query} \\
    \hline

    Who is the enclave within of Montreal? &
   \texttt{select distinct ?sbj where \{ ?sbj wdt:P501 wd:Q340 . ?sbj wdt:P31 wd:Q171441 \}} &
   \multirow{2}{*}{DFSL} 
   & \texttt{select distinct ?obj where \{ \hlred{wd:Q340 wdt:P501 ?obj} . ?obj wdt:P31 wd:Q171441 \}} \\
   \cline{3-4} 
   &  & 
   DFSL-MQ & 
   \texttt{select distinct ?sbj where \{ \hlgreenforest{?sbj wdt:P501 wd:Q340} . ?sbj wdt:P31 wd:Q171441 \}}  \\
   \hline

   The trachea is of what anatomical branch? &
   \texttt{select distinct ?answer where \{ ?answer wdt:P3261 wd:Q175449\}} &
   \multirow{2}{*}{DFSL} 
   & \texttt{select distinct ?answer where \{ \hlred{wd:Q175449 wdt:P3261 ?answer}\}} \\
   \cline{3-4} 
   &  & 
   DFSL-MQ & 
   \texttt{select distinct ?answer where \{ \hlgreenforest{?answer wdt:P3261 wd:Q175449}\}}  \\
   \hline

   What revolution caused the destruction of the Russian Empire? &
   \texttt{select distinct ?obj where \{ wd:Q34266 wdt:P770 ?obj . ?obj wdt:P31 wd:Q10931 \}} &
   \multirow{2}{*}{DFSL} 
   & \texttt{select distinct ?sbj where \{ \hlred{?sbj wdt:P770 wd:Q34266} . ?sbj wdt:P31 wd:Q10931 \}} \\
   \cline{3-4} 
   &  & 
   DFSL-MQ & 
   \texttt{select distinct ?obj where \{ \hlgreenforest{wd:Q34266 wdt:P770 ?obj} . ?obj wdt:P31 wd:Q10931 \}}  \\
   \hline
    
  \end{tabular}
}
  \caption{Some triple-flip errors that are solved by DFSL-MQ.}\label{tab:DFSL_errors}
\end{table*}

\section{Few-shot Learning Examples}\label{app:fs_examples}
\begin{figure*}[ht]
    \centering
    \includegraphics[scale=0.7]{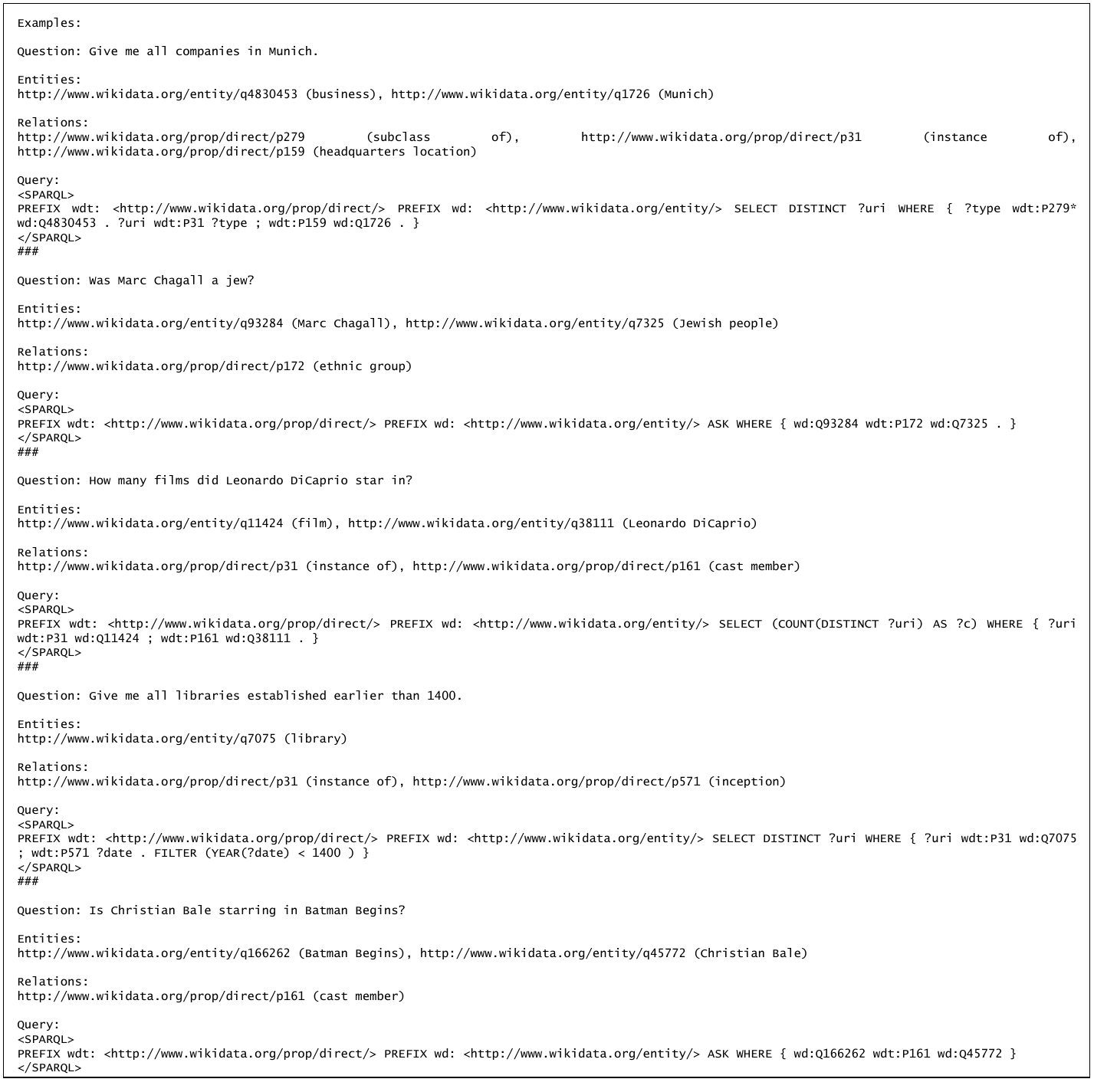}
    \caption{Examples injected in the Few-shot-learning baseline prompt. }\label{fig:appendix_fsl_examples}
\end{figure*}

We report in Figure~\ref{fig:appendix_fsl_examples} the examples selected for the Few-shot learning baseline prompt. The five examples were chosen to be the most representative of the training set, including queries of different kind and structure, such as ASK, COUNT and SELECT. 

\end{document}